\documentclass[runningheads]{llncs}

\usepackage{amsmath,amssymb}
\usepackage{graphicx}
\usepackage{booktabs}
\usepackage{placeins}
\usepackage{subfig}
\usepackage{url}
\usepackage{microtype}
\usepackage[hidelinks]{hyperref}

\title{Modeling Information Blackouts in Missing Not-At-Random Time Series Data}
\titlerunning{Modeling Information Blackouts in MNAR Time Series Data}

\author{Aman Sunesh \and Allan Ma \and Siddarth Nilol}
\authorrunning{A. Sunesh et al.}
\institute{New York University}

\begin{document}
\maketitle

\begin{abstract}
Traffic forecasting systems rely on fixed sensor networks that frequently exhibit contiguous \emph{blackouts}. Such outages are usually treated as ignorable missingness, although dropout can depend on unobserved traffic conditions. We study this possibility with an MNAR-aware latent state-space model that combines linear traffic dynamics with a Bernoulli missingness channel whose probability depends on the latent state. Inference uses an Extended Kalman Filter (EKF) followed by Rauch--Tung--Striebel (RTS) smoothing, and parameters are learned by approximate EM. We evaluate Seattle using a leakage-free, month-balanced set of 300 \emph{unique} all-horizon-aligned blackout windows. On this benchmark, MAR-LDS attains 4.264 mph pooled imputation RMSE and MNAR-LDS improves it to 4.177 (difference $-0.086$); the detector-cluster bootstrap 95\% interval is $[-0.182,-0.002]$. A causal one-step predicted latent representation raises missingness ROC-AUC from 0.685 using observed-only features to 0.784. We further test whether this compact probabilistic model remains competitive with substantially larger neural time-series architectures under the identical masked-imputation protocol. MNAR-LDS ranks second in pooled RMSE and outperforms 8 of 9 evaluated neural architectures; it is within 1.22\% of the best neural result, with no statistically resolved difference under detector-cluster bootstrap, while achieving lower P95 error, lower long-blackout RMSE, and orders of magnitude fewer stored scalar entries. MNAR roughly doubles end-to-end training time relative to MAR and increases EKF+RTS inference time by 41\%, making the accuracy--complexity--cost tradeoff explicit. Controlled state-dependent blackouts further show larger gains when dropout is genuinely informative, including a 6.34\% reduction in 30-minute forecast RMSE relative to MAR.
\end{abstract}

\section{Introduction}
\label{sec:intro}

Traffic forecasting systems rely on dense networks of fixed sensors such as inductive loop detectors to provide continuous measurements of traffic speed and flow. In deployed systems, sensors frequently experience \emph{blackouts}: contiguous intervals of missing readings lasting from minutes to hours. These outages arise from hardware faults, maintenance, and communication failures, and they create systematic gaps that degrade both reconstruction during the blackout and prediction immediately after recovery.

Most pipelines handle missing values implicitly---for example, by masking missing measurements, skipping updates when readings are absent, or applying interpolation. These practices treat the missingness process as effectively \emph{ignorable}. Under Missing At Random (MAR),
\begin{equation}
P(m_t \mid x_t^{(\mathrm{obs})},x_t^{(\mathrm{mis})})
=
P(m_t \mid x_t^{(\mathrm{obs})}),
\end{equation}
where $m_t$ is a missingness indicator and $x_t=(x_t^{(\mathrm{obs})},x_t^{(\mathrm{mis})})$. Under MAR, together with the usual parameter-distinctness and regularity conditions, the missingness mechanism is ignorable for likelihood-based inference~\cite{rubin1976inference,little2002statistical}.

Traffic blackouts can violate this assumption. Congestion, abnormal flow regimes, network stress, or correlated operating conditions may affect both the latent traffic state and the probability of sensor or communication failure. This motivates a Missing Not At Random (MNAR) formulation,
\begin{equation}
P(m_t \mid z_t,x_t^{(\mathrm{obs})})
\neq
P(m_t \mid x_t^{(\mathrm{obs})}),
\end{equation}
where $z_t$ is an unobserved traffic state. If the blackout pattern is informative about $z_t$, then missingness itself can act as an observation rather than only as a preprocessing nuisance~\cite{rubin1976inference,little2002statistical,diggle1994informative}.

Our central question is therefore:
\emph{when detectors exhibit contiguous blackouts, does explicitly modeling blackout events as state-dependent improve reconstruction during the outage and short-horizon prediction after the outage beyond what is obtained from temporal dynamics alone?}

We model traffic with a low-dimensional latent state $z_t\in\mathbb{R}^K$,
\begin{equation}
z_t\mid z_{t-1}\sim\mathcal{N}(Az_{t-1},Q),
\end{equation}
and measurements $x_t\in\mathbb{R}^D$,
\begin{equation}
x_t\mid z_t\sim\mathcal{N}(Cz_t,R).
\end{equation}
For detector $d$, the general missingness model is
\begin{equation}
\Pr(m_{t,d}=1\mid z_t,f_t,g_d)
=
\sigma(b_d+\phi_d^\top z_t+\psi_d^\top f_t+\eta^\top g_d),
\label{eq:mnar_intro}
\end{equation}
where $f_t$ and $g_d$ denote optional temporal and detector covariates, and the detector-feature coefficient $\eta$ is shared across detectors. In the reported Seattle experiments, the active channel is intentionally restricted to the detector intercept and latent-state term, $b_d+\phi_d^\top z_t$; the additional covariates are retained in the formulation for extensibility but are not used to generate the reported Seattle gain.

The resulting model is nonlinear only through the Bernoulli-logistic missingness channel. We therefore use an EKF update with an analytic Jacobian, followed by RTS smoothing. The key advantage of this choice is that it preserves the compact state-space computation and the exact linear-Gaussian speed observation model while adding a smooth local approximation only where the missingness channel is nonlinear. Throughout, we use \emph{causal} in the time-series filtering sense: a representation at time $t$ depends only on information available up to time $t$ and not on future observations; this does not denote causal-effect identification.

Our contributions are fourfold. First, we formulate contiguous traffic-sensor blackouts as an informative-missingness problem by augmenting a compact latent dynamical system with a state-dependent Bernoulli observation channel and approximate EKF--RTS inference. Second, we evaluate whether this compact probabilistic model can compete with substantially larger neural time-series architectures under an identical leakage-free masked-imputation protocol, rather than comparing only against simple interpolation baselines. Third, we provide predictive and statistical evidence for the modeling premise through a causal next-step missingness diagnostic and paired-window, month-stratified, and detector-cluster bootstrap analysis. Finally, we quantify computational cost and development-pool sensitivity and use controlled state-dependent blackouts on METR-LA to characterize when explicit MNAR modeling is useful.

Temporal dynamics account for most of the improvement over naive imputation. Explicit MNAR modeling provides a smaller but statistically supported Seattle imputation refinement, while the accuracy--complexity comparison shows that a compact state-space model can remain competitive with far larger neural architectures. Its value becomes clearer in controlled settings where state-dependent dropout is stronger.

\section{Related Work}
\label{sec:related}

\subsection{Traffic forecasting and state-space inference}
Traffic state estimation has long used latent dynamical systems and Kalman-style inference, with missing measurements commonly handled by skipping the measurement update~\cite{wang2005ekf}. This is computationally attractive but treats the absence of a reading as uninformative. Modern traffic forecasters instead use recurrent, convolutional, or graph neural models to learn temporal and spatial structure~\cite{yu2018stgcn}; these models are powerful predictors but typically handle missingness by preprocessing, masks, or learned representations rather than by specifying a probabilistic mechanism for why observations disappear.

\subsection{Time-series imputation beyond simple interpolation}
Time-series imputation spans several model families. Recurrent approaches such as GRU-D incorporate masks and time-since-observation through learned decay dynamics~\cite{che2018grud}. Generative approaches include GAN-based multivariate imputation~\cite{luo2018gan}, recurrent denoising autoencoders~\cite{zhang2019rdae}, and probabilistic latent-variable models such as GP-VAE~\cite{fortuin2020gpvae}. Self-attention and Transformer approaches explicitly model long-range temporal and cross-variable dependencies, including SAITS~\cite{du2023saits}, Transformer-based imputation~\cite{yildiz2022transformer}, TimesNet~\cite{wu2023timesnet}, and PatchTST-style representations~\cite{nie2023patchtst}. Graph-structured models such as StemGNN learn multivariate dependencies jointly with temporal structure~\cite{cao2020stemgnn}. More traditional machine-learning strategies have also been used for incomplete time series, including SVM-based forecasting~\cite{wu2015missing} and sequential random-forest imputation~\cite{mital2020rf}.

These methods motivate a stronger empirical question than comparison against LOCF alone: can a compact explicit MNAR state-space model remain competitive with expressive neural imputers, including architectures such as TimesNet that are orders of magnitude larger in stored scalar count, when trained and evaluated under the same held-out blackout protocol? We address this accuracy--complexity comparison directly in Sec.~\ref{sec:deep_baselines}. Importantly, strong imputation does not by itself identify an MNAR mechanism. Our contribution is complementary: we ask both whether the missingness indicator is informative about a latent dynamical state and whether explicitly modeling that channel yields a competitive low-complexity alternative to much larger predictive models.

\subsection{MNAR modeling and informative missingness}
In classical missing-data theory, MNAR mechanisms arise when missingness depends on unobserved variables, requiring an explicit model of the missingness process or a shared latent representation~\cite{rubin1976inference,little2002statistical,diggle1994informative}. Despite this long statistical literature, explicit informative-missingness models are less common in traffic state estimation. We bridge these views by augmenting a compact traffic LDS with a detector-wise Bernoulli missingness channel linked to the latent state.

\section{Methods}
\label{sec:methods}

\subsection{Latent state-space model}
Let $z_t\in\mathbb{R}^K$ denote a latent traffic state at 5-minute interval $t$, and $x_t\in\mathbb{R}^D$ the detector speed vector:
\begin{align}
z_t\mid z_{t-1} &\sim \mathcal{N}(Az_{t-1},Q),\\
x_t\mid z_t &\sim \mathcal{N}(Cz_t,R).
\end{align}
Here $A,Q\in\mathbb{R}^{K\times K}$, $C\in\mathbb{R}^{D\times K}$, and $R$ is constrained to be diagonal. The MAR baseline uses the same LDS but ignores the missingness mask as an observation channel.

\subsection{Latent-state-dependent missingness}
Let $m_{t,d}\in\{0,1\}$ denote whether detector $d$ is missing at time $t$. The general model is
\begin{equation}
\Pr(m_{t,d}=1\mid z_t,f_t,g_d)
=
\sigma\!\left(b_d+\phi_d^\top z_t+\psi_d^\top f_t+\eta^\top g_d\right).
\label{eq:mnar_logistic}
\end{equation}
The Seattle experiments activate only $b_d+\phi_d^\top z_t$. Thus the MNAR result is not driven by additional time-of-day or detector metadata. Artificially inserted evaluation masks are tracked separately and excluded from missingness-model updates, so the model cannot learn the held-out evaluation pattern as if it were naturally occurring dropout.

\subsection{EKF approximation and why we use it}
The speed observation model is linear-Gaussian, but the sigmoid in~\eqref{eq:mnar_logistic} makes exact filtering intractable. Let
\begin{align}
\ell_{t,d} &= b_d+\phi_d^\top \mu_{t|t-1},\\
\pi_{t,d} &= \sigma(\ell_{t,d}).
\end{align}
Around the predicted state mean, the Bernoulli channel is locally Gaussianized:
\begin{equation}
m_{t,d}\approx \pi_{t,d}+\epsilon_{t,d},\qquad
\epsilon_{t,d}\sim\mathcal{N}(0,s_{t,d}).
\end{equation}
For the experiments, we use the moment-matched Bernoulli variance
\begin{equation}
s_{t,d}=\pi_{t,d}(1-\pi_{t,d}),
\end{equation}
while a constant-variance alternative is evaluated as a sensitivity check. The analytic Jacobian is
\begin{equation}
J_{t,d}
=
\frac{\partial \pi_{t,d}}{\partial z_t}
=
\pi_{t,d}(1-\pi_{t,d})\phi_d.
\end{equation}
We then combine the observed speed likelihood and the Gaussianized missingness likelihood in an EKF-style information update.

The EKF is a pragmatic choice rather than a claim of globally optimal approximate inference. The nonlinearity is smooth and confined to the logistic missingness channel, the Jacobian is analytic, and the latent state is low-dimensional. This preserves the state-space structure and requires only $K\times K$ linear solves after detector-wise accumulation, instead of introducing particles or a second amortized inference model. As an approximation sensitivity check, replacing the moment-matched missingness variance with a constant variance changes the reported RMSEs by less than 0.003 mph across the evaluated horizons.

\subsection{RTS smoothing, approximate EM, and forecasting}
After filtering, we apply standard RTS smoothing:
\begin{align}
G_t &= \Sigma_{t|t}A^\top\Sigma_{t+1|t}^{-1},\\
\mu_{t|T} &= \mu_{t|t}+G_t(\mu_{t+1|T}-\mu_{t+1|t}),\\
\Sigma_{t|T}
&=
\Sigma_{t|t}
+
G_t(\Sigma_{t+1|T}-\Sigma_{t+1|t})G_t^\top.
\end{align}
The approximate EM E-step runs EKF+RTS. The LDS M-step updates $(\mu_0,\Sigma_0,A,Q,C,R)$ from smoothed moments using only observed speeds. Lag-one posterior cross-moments required by the LDS M-step are approximated using smoothed-state statistics rather than an exact RTS-EM cross-covariance recursion. To stabilize EM, we regularize the dynamics by shrinking $A$ toward the identity and $Q$ toward an isotropic covariance, with an upper bound on $\mathrm{tr}(Q)$; the same stabilization is used for MAR and MNAR and is not tuned on the 300-window evaluation set.

The missingness parameters are updated detector-wise by maximizing
\begin{equation}
\mathcal{L}_d
=
\sum_{t\in\mathcal{T}_d}
\left[
m_{t,d}\log\pi_{t,d}
+
(1-m_{t,d})\log(1-\pi_{t,d})
\right],
\end{equation}
where $\mathcal{T}_d$ excludes artificially held-out evaluation cells.

The reference protocol fixes $K=20$, trains MAR for 10 EM iterations, initializes MNAR from the resulting MAR-10 checkpoint with missingness parameters reset, and performs 10 additional MNAR EM iterations. Missingness parameters receive two gradient steps per EM iteration with learning rate $10^{-4}$. We use this pre-specified $10+10$ protocol without selecting EM depth on the 300-window evaluation set. 

To control for unequal optimization depth, we extended MAR beyond 10 EM iterations on the disjoint development pool. Additional MAR optimization produced only marginal improvements: RMSE was 4.992 at iteration 10, 4.988 at iteration 20, and 4.984 at iteration 30. Thus, 20 additional MAR iterations reduced development-pool RMSE by only 0.008 mph; this small change, relative to the 0.0865 mph Seattle MAR-to-MNAR difference, argues against additional EM updates alone explaining the MNAR refinement.

For imputation we use $\hat{x}_t=C\mu_{t|T}$. For a $k$-step forecast after blackout termination at $b$,
\begin{align}
\mu_{b+k|b} &= A^k\mu_{b|b},\\
\Sigma_{b+k|b}
&=
A^k\Sigma_{b|b}(A^\top)^k
+
\sum_{i=0}^{k-1}A^iQ(A^\top)^i,
\end{align}
and map the state through $C$.

\section{Results}
\label{sec:results}

\subsection{Datasets and evaluation protocol}
Our primary real-data evaluation uses the Seattle Inductive Loop Detector dataset (2015)~\cite{seattleloop2015}, containing 5-minute traffic speed measurements across 147 detectors over one year with roughly 5\% missing readings. We evaluate (i) reconstruction inside a blackout and (ii) post-blackout point forecasting at 1, 3, and 6 steps (5, 15, and 30 minutes).

We first form a 441-window pool for which the same \texttt{window\_id} has an imputation target and valid 1-, 3-, and 6-step forecast targets. From this pool, we select \textbf{300 unique} window IDs using deterministic month-balanced sampling without replacement. Each month receives up to 25 windows initially; any deficit caused by months with fewer available aligned windows is distributed across months with unused aligned windows. Selection uses no model performance information. This removes duplicate sampled entries while preserving temporal coverage and all-horizon alignment.

For every selected blackout, the target detector segment is removed from the input panel before model fitting or inference.  Artificial evaluation masks are excluded from both missingness-parameter learning and the MNAR missingness-observation channel during inference; they are used only to withhold the corresponding speed measurements. Naturally occurring missingness remains available to the MNAR observation channel.

Naturally missing values remain as missing inputs but are never evaluation targets. The 300-window evaluation set contains 4,236 event target points (4,188 unique physical detector-time cells because a small number of event intervals overlap).

The neural comparisons in Sec.~\ref{sec:deep_baselines} use the same 300 blackout events and the same 4,236 event target points. Code, frozen evaluation-window IDs, and experiment artifacts are available in the \href{https://github.com/BlackoutBayes/Modeling-Information-Blackouts-in-MNAR-Time-Series}{project repository}. Neural training windows overlapping any evaluation blackout are excluded, and validation masks are generated only on training-side sequence windows. Hyperparameters are fixed without using the 300-window evaluation set for model selection.

\subsection{Seattle LDS results}
\label{sec:seattle_results}

Table~\ref{tab:seattle_rmse_main} reports results on the unique-300 evaluation set. Latent dynamics account for the dominant gain, while the MNAR channel provides a smaller refinement.

\begin{table}[t]
\centering
\caption{Seattle Loop (2015), unique-300 protocol: pooled RMSE (mph). Lower is better.}
\label{tab:seattle_rmse_main}
\begin{tabular}{lcccc}
\toprule
\textbf{Method} & \textbf{Impute} & \textbf{1-step} & \textbf{3-step} & \textbf{6-step}\\
\midrule
LOCF & 7.364 & 8.132 & 8.783 & 9.075\\
LinearInterp + SeasonalNaive & 4.974 & 8.583 & 9.064 & 9.232\\
\midrule
MAR (LDS) & 4.264 & 4.183 & 4.509 & 4.517\\
MNAR (LDS) & \textbf{4.177} & \textbf{4.124} & \textbf{4.434} & \textbf{4.497}\\
\bottomrule
\end{tabular}
\end{table}
Moving from LOCF to MAR reduces imputation MSE by 66.5\%. The additional MAR-to-MNAR imputation reduction is 4.0\% in MSE, corresponding to a pooled RMSE change of $-0.0865$ mph; the detector-cluster bootstrap 95\% CI for this difference is $[-0.1823,-0.0016]$ mph. Full paired, month-stratified, and detector-cluster uncertainty estimates are reported in Table~\ref{tab:bootstrap}. Forecast point estimates also favor MNAR at all three horizons ($-0.0589$, $-0.0751$, and $-0.0203$ mph for 1, 3, and 6 steps), but we interpret these as modest point-estimate refinements rather than strong claims of forecast superiority.

\begin{figure}[!htbp]
    \centering
    \includegraphics[width=\linewidth]{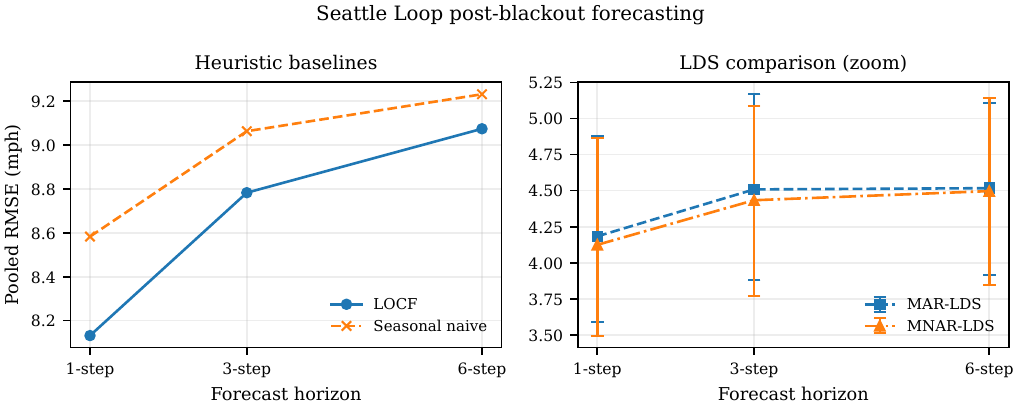}
    \caption{Seattle post-blackout forecasting under the unique-300 protocol. The left panel shows heuristic baselines; the right panel zooms the MAR-LDS/MNAR-LDS comparison with 95\% bootstrap confidence intervals.}
    \label{fig:forecast}
\end{figure}

\subsection{Bootstrap robustness of the MNAR--MAR gain}
\label{sec:bootstrap}

Because the Seattle effect is small, ordinary iid uncertainty alone is not sufficient. We therefore pair MAR and MNAR by the exact evaluation occurrence and report both equal-window and pooled-RMSE estimators under several resampling schemes. The pooled difference is
\begin{equation}
\Delta_{\mathrm{pool}}
=
\mathrm{RMSE}_{\mathrm{MNAR}}
-
\mathrm{RMSE}_{\mathrm{MAR}}
=
-0.0865\ \mathrm{mph}.
\end{equation}
The mean paired per-window RMSE difference is $-0.0907$ mph.

\begin{table}[t]
\centering
\caption{Bootstrap uncertainty for the Seattle MNAR--MAR imputation difference. Negative values favor MNAR; all intervals are 95\% confidence intervals.}
\label{tab:bootstrap}
\begin{tabular}{lcc}
\toprule
\textbf{Resampling} & $\boldsymbol{\Delta}$ \textbf{pooled RMSE} & \textbf{95\% CI}\\
\midrule
Paired window & -0.0865 & [-0.1696,\,-0.0081]\\
Month-stratified & -0.0865 & [-0.1685,\,-0.0099]\\
Detector-cluster & -0.0865 & [-0.1823,\,-0.0016]\\
\bottomrule
\end{tabular}
\end{table}

All three pooled-RMSE intervals remain below zero. The same conclusion holds for the equal-window estimator: paired-window, month-stratified, and detector-cluster intervals are respectively $[-0.1450,-0.0392]$, $[-0.1447,-0.0414]$, and $[-0.1589,-0.0302]$. Thus the Seattle gain is small, but it is not explained solely by treating correlated windows as iid observations.

\begin{figure}[!htbp]
    \centering
    \includegraphics[width=0.88\linewidth]{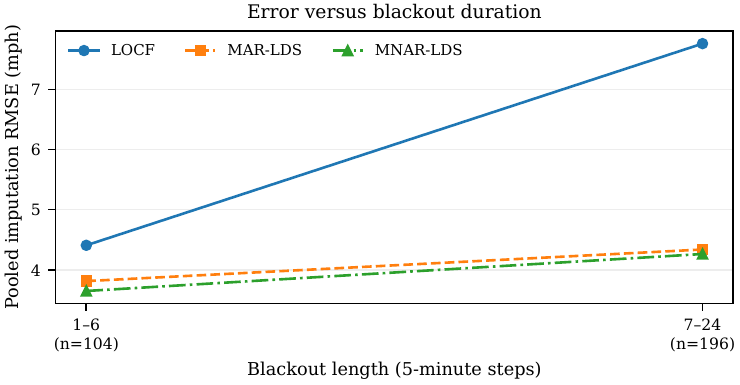}
    \caption{Pooled imputation RMSE by nonempty blackout-length bucket. Distinct markers and line styles supplement color, and bucket counts are shown explicitly.}
    \label{fig:length}
\end{figure}

\subsection{Modern neural baselines under masked imputation}
\label{sec:deep_baselines}

To compare against stronger predictive alternatives, we evaluate native time-series imputation models and general time-series architectures adapted to the same masked-imputation protocol covering mask/decay recurrence (GRU-D), self-attention (SAITS), task-general temporal modeling (TimesNet), patch Transformers (PatchTST), graph-structured multivariate modeling (StemGNN), frequency-domain and convolutional models (FreTS and TCN), a simple decomposition-linear model (DLinear), and probabilistic latent-variable imputation (GP-VAE). All models use the same 300 events, identical held-out targets, leakage checks, and early stopping. StemGNN learns inter-series correlations from the data; it is graph-structured but does not use a physical Seattle road-adjacency graph.

\begin{table}[t]
\centering
\caption{Seattle imputation on the identical 300-window protocol (4,236 event target points). Pooled RMSE weights target cells equally; mean-window RMSE weights events equally. Lower is better.}
\label{tab:deep_baselines}
\small
\begin{tabular}{lccc}
\toprule
\textbf{Method} & \textbf{Pooled RMSE} & \textbf{Mean-window RMSE} & \textbf{MAE}\\
\midrule
TimesNet~\cite{wu2023timesnet} & \textbf{4.126} & 3.047 & 2.670\\
MNAR-LDS & 4.177 & 3.243 & 2.835\\
SAITS~\cite{du2023saits} & 4.242 & \textbf{3.026} & \textbf{2.610}\\
MAR-LDS & 4.264 & 3.334 & 2.910\\
PatchTST~\cite{nie2023patchtst} & 4.287 & 3.195 & 2.783\\
FreTS~\cite{yi2023frets} & 4.422 & 3.387 & 2.938\\
StemGNN~\cite{cao2020stemgnn} & 4.436 & 3.348 & 2.898\\
TCN~\cite{bai2018tcn} & 4.452 & 3.238 & 2.779\\
DLinear~\cite{zeng2023dlinear} & 4.798 & 3.399 & 3.004\\
GRU-D~\cite{che2018grud} & 4.874 & 3.433 & 3.074\\
LinearInterp & 4.974 & 3.206 & 2.905\\
GP-VAE~\cite{fortuin2020gpvae} & 5.469 & 3.734 & 3.439\\
LOCF & 7.364 & 3.991 & 3.794\\
\bottomrule
\end{tabular}
\end{table}

Table~\ref{tab:deep_baselines} shows that MNAR-LDS ranks second in pooled RMSE, outperforming 8 of 9 evaluated neural architectures. TimesNet obtains the lowest pooled RMSE, but its numerical advantage over MNAR-LDS is only 0.0510 mph (1.22\%). Under detector-cluster resampling, the 95\% interval for $\mathrm{RMSE}_{\mathrm{TimesNet}}-\mathrm{RMSE}_{\mathrm{MNAR}}$ is $[-0.332,+0.267]$, so this ordering is not statistically resolved. SAITS, meanwhile, has the best event-macro RMSE and MAE. Hence no single method dominates every summary.

The comparison is also informative about tail behavior and model size. On the same 300 events, MNAR-LDS has lower P95 per-window RMSE (8.245 versus 8.634 for TimesNet) and lower pooled RMSE on the top-quartile long-blackout subset (4.383 versus 4.404; blackout length at least 24 steps). TimesNet contains 37,576,851 trainable parameters, whereas the stored MNAR parameter structure contains 28,856 numeric entries; the corresponding raw parameter storage is approximately 143.3 MiB versus 0.220 MiB.

\begin{table}[t]
\centering
\caption{Accuracy--complexity comparison with the strongest pooled-RMSE neural baseline. The scalar-count row is a storage-complexity comparison, not an assertion that every stored LDS entry is an independent degree of freedom.}
\label{tab:timesnet_tradeoff}
\begin{tabular}{lcc}
\toprule
\textbf{Quantity} & \textbf{MNAR-LDS} & \textbf{TimesNet}\\
\midrule
Pooled RMSE & 4.177 & \textbf{4.126}\\
P95 window RMSE & \textbf{8.245} & 8.634\\
Long-blackout RMSE & \textbf{4.383} & 4.404\\
Stored/trainable scalar entries & \textbf{28,856} & 37,576,851\\
Raw parameter storage & \textbf{0.220 MiB} & 143.344 MiB\\
\bottomrule
\end{tabular}
\end{table}

We do not interpret wall-clock fit time across MNAR and TimesNet as an architecture-intrinsic speed comparison because the measured MNAR fit ran on CPU whereas TimesNet trained on an NVIDIA A100. The size comparison, in contrast, is hardware independent.

\subsection{Causal evidence that the missingness pattern contains latent-state information}
\label{sec:missingness_diagnostic}

Because RTS smoothing uses future observations, an RTS-smoothed latent representation is descriptive but acausal. We therefore evaluate both causal and acausal representations on a chronological split. The latent representations in this diagnostic are obtained from the MAR-LDS, which does not use the missingness indicator as an observation channel, and the one-step predicted state is computed without access to future observations. 

Table~\ref{tab:auc} compares observed proxies with the filtered state and the one-step predicted latent state, while the smoothed representation is included only as an explicitly acausal reference.

\begin{table}[t]
\centering
\caption{Next-step missingness prediction diagnostic. The predicted state is causal and uses no future observations.}
\label{tab:auc}
\begin{tabular}{lccc}
\toprule
\textbf{Feature representation} & \textbf{ROC-AUC} & \textbf{AP} & \textbf{Causal?}\\
\midrule
Observed-only proxies & 0.6849 & 0.4672 & yes\\
Filtered latent state & 0.7844 & 0.5545 & yes\\
One-step predicted latent state & \textbf{0.7844} & \textbf{0.5546} & yes\\
RTS-smoothed latent state & 0.7831 & 0.5540 & no\\
\bottomrule
\end{tabular}
\end{table}

\begin{figure}[!htbp]
    \centering
    \includegraphics[width=0.78\linewidth]{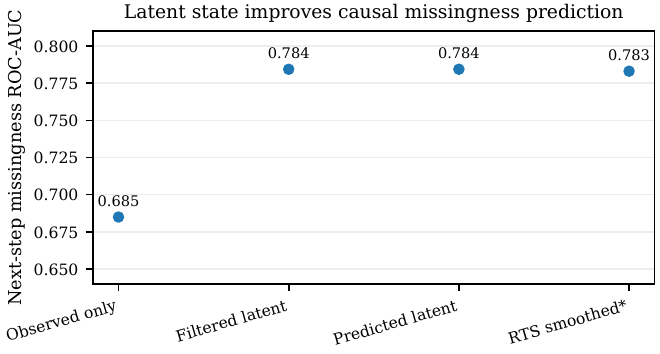}
    \caption{Next-step missingness ROC-AUC for observed-only features, causal latent representations, and the acausal RTS-smoothed reference. The causal predicted latent state improves substantially over observed-only features.}
    \label{fig:missingness_auc}
\end{figure}

As visualized in Fig.~\ref{fig:missingness_auc}, the causal predicted latent representation improves ROC-AUC over observed-only features by $+0.0995$, with bootstrap 95\% CI $[+0.0824,+0.1172]$. The acausal smoother does not outperform the causal predicted state. This does not prove that every real blackout is MNAR, but it is substantially stronger evidence for the modeling premise than an acausal smooth-state diagnostic: latent traffic dynamics contain information about imminent missingness that is not captured by the observed proxies alone.

\subsection{Computational cost of adding the MNAR channel}
\label{sec:cost}

The practical value of a small error reduction depends on its cost. We therefore time the reference fit and repeat EKF+RTS inference five times on the same masked Seattle panel.

\begin{table}[t]
\centering
\caption{Computational cost of the Seattle LDS models on the same CPU environment. MNAR training is warm-started from MAR, so its end-to-end fit includes both phases.}
\label{tab:cost}
\begin{tabular}{lcc}
\toprule
\textbf{Quantity} & \textbf{MAR} & \textbf{MNAR}\\
\midrule
Base MAR fit & 633.78 s & 633.78 s\\
Additional MNAR refinement & -- & 664.64 s\\
End-to-end fit & 633.78 s & 1298.42 s\\
EKF+RTS inference & $14.73\pm0.67$ s & $20.83\pm0.30$ s\\
Inference peak RSS & 1787.7 MB & 1785.5 MB\\
\bottomrule
\end{tabular}
\end{table}

The MNAR refinement costs 1.05$\times$ the MAR fit itself, making the end-to-end MNAR training pipeline 2.05$\times$ as long. MNAR inference is 1.41$\times$ slower. Peak process memory is essentially unchanged. The practical tradeoff is therefore primarily additional computation rather than additional memory.

\subsection{Latent-dimension and approximation sensitivity}
\label{sec:sensitivity}

Latent-dimension sensitivity experiments use a separate 39-window aligned development pool that is disjoint from the 300-window evaluation set. We retain $K=20$ in the primary model because it is the pre-specified reference dimension; the evaluation set is not used for latent-dimension selection.

\begin{table}[t]
\centering
\caption{Development-pool sensitivity to latent dimension. The 39 windows are disjoint from the 300-window evaluation set.}
\label{tab:k_sensitivity}
\begin{tabular}{cccc}
\toprule
$K$ & \textbf{MAR RMSE} & \textbf{MNAR RMSE} & \textbf{MNAR--MAR}\\
\midrule
10 & 5.6270 & 5.7642 & +0.1372\\
20 & 4.9924 & 4.9965 & +0.0041\\
30 & 4.8610 & 4.8133 & -0.0477\\
40 & 4.7177 & \textbf{4.5469} & \textbf{-0.1709}\\
\bottomrule
\end{tabular}
\end{table}

Absolute development error decreases for both models as capacity increases over the tested range. The MNAR--MAR difference also changes with latent dimension: MNAR is worse at $K=10$, approximately tied with MAR at $K=20$, and increasingly favorable at $K=30$ and $K=40$. Thus, we do \emph{not} claim that $K=20$ is globally optimal. The model retains the pre-specified $K=20$ configuration rather than selecting the latent dimension that maximizes the MNAR--MAR gap on the development pool. The missingness-variance approximation is also insensitive: switching between moment-matched and constant variance changes RMSE by less than 0.003 mph across the evaluated tasks.

On the disjoint development pool, missingness weights from $0.25$ to $1.0$ produce pooled imputation RMSEs within 0.008 mph of one another (4.989--4.996), while increasing the weight to $2.0$ degrades RMSE to 5.016, indicating robustness over moderate weighting but sensitivity to over-weighting the missingness channel.

\begin{figure}[!htbp]
    \centering
    \includegraphics[width=0.85\linewidth]{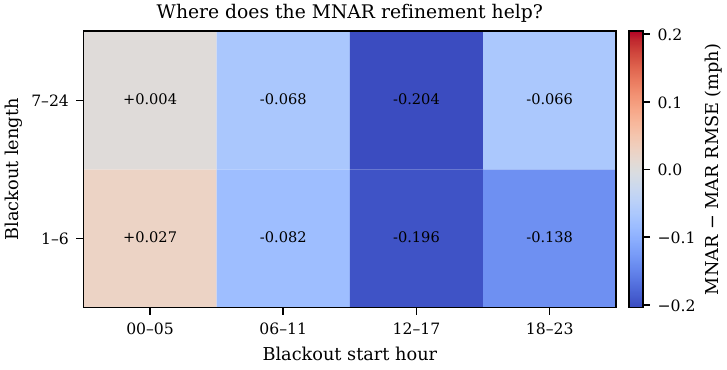}
    \caption{Mean paired $\Delta$RMSE $=$ RMSE$_{\mathrm{MNAR}}-$RMSE$_{\mathrm{MAR}}$ by blackout-length and start-hour bucket. Negative values favor MNAR. A single zero-centered difference map avoids comparing separate color scales.}
    \label{fig:delta_heatmap}
\end{figure}

\subsection{Controlled state-dependent blackouts on METR-LA}
\label{sec:metrla}

Seattle contains naturally occurring blackouts but does not provide ground truth for their missingness mechanism. We therefore use the METR-LA traffic dataset~\cite{li2018dcrnn} in a controlled experiment in which blackout occurrence is injected as a function of the underlying traffic state.

\begin{table}[t]
\centering
\caption{METR-LA with synthetic state-dependent blackouts: RMSE on held-out blackout windows. Lower is better.}
\label{tab:metrla}
\begin{tabular}{lcccc}
\toprule
\textbf{Method} & \textbf{Impute} & \textbf{1-step} & \textbf{3-step} & \textbf{6-step}\\
\midrule
LOCF & 9.433 & 9.855 & 9.262 & 9.147\\
MAR (LDS) & 5.483 & 4.700 & 5.320 & 5.114\\
MNAR (LDS) & \textbf{5.355} & \textbf{4.700} & \textbf{5.238} & \textbf{4.790}\\
\bottomrule
\end{tabular}
\end{table}

Under controlled state-dependent dropout, MNAR reduces imputation RMSE from 5.483 to 5.355, a 2.33\% reduction relative to MAR. The advantage becomes larger at the longest forecast horizon: 6-step RMSE falls from 5.114 to 4.790, a 6.34\% reduction, while 3-step RMSE improves by 1.54\%. The 1-step difference is negligible. These results reinforce our main interpretation: adding a missingness channel is not universally beneficial, but can provide a materially larger gain when dropout is genuinely coupled to the underlying latent regime.

We further sweep the true state-dependence strength $\alpha$. At $\alpha=0$, the MNAR--MAR difference is approximately $-0.011$ RMSE. At $\alpha=0.4$ it is $-0.227$, at $\alpha=0.8$ it is $-0.052$, and at $\alpha=1.2$ it is $-0.190$. The trend is therefore favorable but noisy and \emph{not monotonic}; variability also grows at high dependence.

\begin{table}[t]
\centering
\caption{Synthetic MNAR-strength sweep: RMSE mean $\pm$ std across random seeds. Negative $\Delta$ favors MNAR.}
\label{tab:alpha}
\begin{tabular}{cccc}
\toprule
$\alpha$ & \textbf{MAR RMSE} & \textbf{MNAR RMSE} & $\boldsymbol{\Delta}$\textbf{RMSE}\\
\midrule
0.0 & $3.741\pm0.026$ & $3.730\pm0.023$ & $-0.011\pm0.007$\\
0.4 & $3.858\pm0.082$ & $3.630\pm0.063$ & $-0.227\pm0.033$\\
0.8 & $3.750\pm0.445$ & $3.698\pm0.538$ & $-0.052\pm0.109$\\
1.2 & $3.849\pm0.510$ & $3.659\pm0.946$ & $-0.190\pm0.372$\\
\bottomrule
\end{tabular}
\end{table}

\begin{figure}[!htbp]
    \centering
    \includegraphics[width=0.78\linewidth]{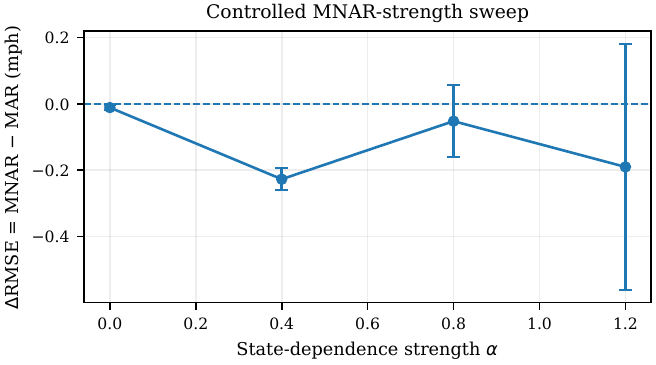}
    \caption{Controlled state-dependence sweep. $\Delta$RMSE denotes MNAR minus MAR, so negative values favor MNAR. The advantage is generally larger when state dependence is present, although the trend is noisy rather than strictly monotonic.}
    \label{fig:alpha}
\end{figure}

\section{Discussion and Limitations}
\label{sec:discussion}

The experiments support a deliberately scoped claim. First, the central mechanism comparison is MAR-LDS versus MNAR-LDS, not LDS versus LOCF. On Seattle, the MNAR improvement is numerically modest but remains negative under paired, month-stratified, and detector-cluster bootstrap resampling. On the disjoint development pool, extending MAR to 20 and 30 EM iterations changes imputation RMSE only marginally, arguing against additional optimization depth alone explaining the refinement. Second, causal missingness prediction provides independent support for the premise that the latent traffic trajectory contains information about imminent dropout. Third, the neural comparison establishes a separate accuracy--complexity result: TimesNet has the best pooled RMSE and SAITS the best event-macro summaries, but the much smaller MNAR-LDS ranks second in pooled RMSE while offering favorable tail and long-blackout behavior. The contribution is therefore not universal neural-model superiority, but evidence that an explicit probabilistic treatment of informative missingness can remain competitive without requiring a large predictive architecture.

The cost analysis also matters. MNAR is not ``free'': its warm-started refinement approximately doubles end-to-end training time and increases inference time by about 41\% relative to MAR. The model is most compelling when a practitioner values explicit missingness semantics, causal latent-state diagnostics, compact storage, or robustness to long outages enough to justify that additional runtime.

Two dataset limitations remain. Seattle provides real blackouts but no ground-truth causal missingness mechanism; METR-LA provides controlled state-dependent missingness but is synthetic with respect to dropout. We therefore do not claim universal generalization across traffic networks. Additional naturally occurring blackout datasets and controlled mechanisms on further sensor networks would be valuable future work.

Methodologically, the LDS is linear-Gaussian and the missingness channel is logistic. EKF local linearization can be inaccurate under highly nonlinear posterior geometry, and the current model does not incorporate incidents, weather, maintenance records, or detector-health covariates that could explain dropout directly. Future work should evaluate nonlinear or graph-structured state-space dynamics, richer missingness channels, and variational or sigma-point inference while retaining explicit separation between the observation process and the missingness mechanism.

\section{Conclusion}
\label{sec:conclusion}

We studied contiguous traffic-sensor blackouts through an explicit informative-missingness model. Under a leakage-free Seattle protocol with 300 unique all-horizon-aligned blackout windows, MAR-LDS achieves 4.264 mph pooled imputation RMSE and MNAR-LDS improves it to 4.177. The improvement is small in absolute terms but survives detector-cluster bootstrap uncertainty. A causal one-step predicted latent state improves missingness ROC-AUC from 0.685 to 0.784 relative to observed-only features, supporting the premise that the latent traffic state contains information about future dropout.

Stronger neural comparisons change the framing but not the mechanism result. TimesNet achieves 4.126 pooled RMSE, 1.22\% below MNAR-LDS, yet the difference is not statistically resolved under detector-cluster resampling; MNAR-LDS also has lower P95 and long-blackout error and uses orders of magnitude fewer stored scalar entries. At the same time, adding the MNAR channel roughly doubles end-to-end training time relative to MAR and raises inference time by about 41\%. The resulting conclusion is therefore a cost-aware one: strong temporal dynamics are the first-order requirement for blackout recovery, while explicit MNAR modeling is an interpretable refinement when the dropout process itself carries information about the latent system state. This effect is modest on naturally occurring Seattle outages but becomes more pronounced under controlled state-dependent dropout, where MNAR reduces 30-minute forecast RMSE by 6.34\% relative to MAR.

\end{document}